\documentclass[double]{article}
\usepackage{times}
\usepackage{algorithm}
\usepackage{algorithmic}
\usepackage{wrapfig}
\usepackage{verbatim}
\usepackage{amsmath}
\usepackage{graphicx}
\usepackage{subcaption}
\usepackage[active]{srcltx}
\usepackage{color}
\usepackage{lineno}
\usepackage{dirtytalk}
\usepackage{amsthm}
\usepackage{authblk}
\usepackage{listings}
\usepackage{tikz}
\usepackage{booktabs}
\usetikzlibrary{shapes.geometric, arrows}

\usepackage{epsfig,graphicx,amsfonts}
\usepackage{amsmath,amsthm,amssymb}
\usepackage{xcolor}

\usepackage{adjustbox}
\usepackage{multirow}
\usepackage{setspace}
\usepackage{hyperref}
\usepackage{comment}

\long\def\comment #1\commentend{}

\begin{document}

\title{A Comprehensive Benchmark of Machine and Deep Learning Across Diverse Tabular Datasets}

\author{Assaf Shmuel$^{1*}$, Oren Glickman$^{1}$, Teddy Lazebnik$^{2,3}$\\
\(^1\) Department of Computer Science, Bar Ilan University, Israel\\
\(^2\) Department of Mathematics, Ariel University, Israel\\
\(^3\) Department of Cancer Biology, Cancer Institute, UCL, UK\\
\(^*\) Corresponding author: assafshmuel91@gmail.com
}

\date{}

\maketitle

\begin{abstract}
The analysis of tabular datasets is highly prevalent both in scientific research and real-world applications of Machine Learning (ML). Unlike many other ML tasks, Deep Learning (DL) models often do not outperform traditional methods in this area. Previous comparative benchmarks have shown that DL performance is frequently equivalent or even inferior to models such as Gradient Boosting Machines (GBMs). 
In this study, we introduce a comprehensive benchmark aimed at better characterizing the types of datasets where DL models excel. 
Although several important benchmarks for tabular datasets already exist, our contribution lies in the variety and depth of our comparison: we evaluate 111 datasets with 20 different models, including both regression and classification tasks. These datasets vary in scale and include both those with and without categorical variables. Importantly, our benchmark contains a sufficient number of datasets where DL models perform best, allowing for a thorough analysis of the conditions under which DL models excel. Building on the results of this benchmark, we train a model that predicts scenarios where DL models outperform alternative methods with 86.1\% accuracy (AUC 0.78). We present insights derived from this characterization and compare these findings to previous benchmarks.

\end{abstract}

\section{Introduction}
\label{sec:introduction}
Machine learning (ML) has long been considered superior to deep learning (DL) when it comes to handling tabular data \cite{why_do,DL_not_all_you_need,tabular_issue}, a common data format in many real-world applications and fields like medicine \cite{intro_bio_1,intro_bio_2,pick_tau_1}, economy \cite{intro_economy_1,intro_economy_2}, and operations \cite{intro_operation_1,intro_operation_2}, to name a few. Nonetheless, this generalization does not hold universally \cite{tabnet}. While traditional ML algorithms, such as Random Forest \cite{rf} and XGboost \cite{xgboost}, often excel in this domain, there are scenarios where DL models outperform ML methods, challenging the prevailing notion \cite{tabnet}. Understanding the conditions under which DL models can surpass ML methods on tabular data is crucial for practitioners seeking to leverage the full potential of these advanced techniques.

Several benchmarking studies have attempted to compare the performance of ML and DL models across various types of data, in general \cite{compare_all_1,compare_all_2,compare_all_3}, and for tabular data, in particular \cite{compare_table}. For instance, \cite{tabnet} proposed the TabNet model, a DL model specifically designed to handle tabular data, and showed competitive performance against traditional ML approaches. \cite{why_do} used 45 tabular datasets from various domains (mainly from OpenML~\cite{OpenML2013}) with heterogeneous columns, below 500 columns but over 3000 rows with at least ten times more rows than columns, no time-series, and without deterministic target column (like poker games' data). The authors removed rows with missing data, used one hot encoding \cite{ohe} for categorical columns, and for regression cases used a log transformation to the target variable. Based on these settings, the authors compared MLP \cite{mlp}, ResNet \cite{resnet}, FT transformer \cite{issue_1}, and SAINT \cite{saint} for the DL models and the RF \cite{rf}, XGboost \cite{xgboost}, histogram-gradient boosting tree \cite{hgbt}, and gradient boosting tree \cite{gbt} for the ML models. The authors show that XGboost outperformed all DL models for both classification and regression tasks while showing that tuning hyperparameters does not make DL models outperform the ML models. They also suggest that DL models are challenged by uninformative features. Similarly, \cite{DL_not_all_you_need} used 11 tabular datasets with both classification and regression problems with 10 to 2000 columns and between 7000 and a million rows. The authors performed standardization (aka z-score normalization) of each feature in each dataset to have a mean of zero and a standard deviation of one. The authors take into consideration the XGboost model as a representer of the ML models, four DL models including TabNet, and three ensemble models of the DL model (with and without XGboost). The authors concluded that the DL models underperform compared to the ML models while an ensemble combining both model types produces better results, on average. \cite{when_do} used 176 classification datasets from OpenML and CC-18 and 19 algorithms including 11 DL models and 8 ML models. The authors found that, on average, CatBoost \cite{catboost} and XGboost outperformed the other models while also being an order of magnitude more computationally effective. The authors used the PyMFE \cite{pymfe} to compute a feature vector for each dataset and used it to analyze the properties in which DL models outperform ML models, on average, finding that irregular, with a large number of rows, and a high ratio of size to number of features actually decrease the DL models performance. \cite{latest_review} compared 300 datasets including both classification and regression tasks from multiple domains further supporting the conclusion that, on average, tree-based ML models outperform DL models. 
 
Despite these efforts, a comprehensive understanding of the nuanced conditions under which DL models excel over ML models, particularly on tabular data, remains underexplored. In particular, measurable (statistical) features of the dataset for the binary prediction of either DL or ML model will provide a better performance, are still unknown. 

In this paper, we aim to fill this gap by conducting an extensive benchmark study involving 111 datasets encompassing both regression and classification tasks. We evaluate 20 different model configurations, including 7 DL-based models, 7 Tree-based Ensemble models (TE), and 6 classical ML-based models, to ascertain their performance on tabular data. Based on these results, we adopted a meta-learning approach, profiling the properties of datasets where DL outperforms ML models. Our results reveal complex patterns while generic behavior like a small number of rows and a large number of columns as well as large kurtosis results in DL models outperforming ML models. We also find that the gap between the two groups is smaller for classification tasks compared to regression tasks. Our key contributions are:

\begin{itemize}
    \item Conducted an extensive and diverse benchmark involving 111 datasets encompassing both regression and classification tasks.
    \item Evaluated the performance of 20 different model configurations, including 7 DL-based models, 7 Tree-based Ensemble models (TE), and 6 classical ML-based models.
    \item Identified dataset characteristics, such as small number of rows and high kurtosis, that favor DL models over other ML models.
    \item Trained a meta-learning model to predict whether DL models will outperform ML models, achieving 86.1\% accuracy (AUC 0.78).
    \item Presented explainable models, namely logistic regression and symbolic regression, which predict where DL models may perform better than alternative models.
    \item Provided detailed insights into the comparative performance of DL and ML models on a diverse set of tabular datasets.
\end{itemize}

The remainder of this paper is structured as follows: Section \ref{sec:experimental_setup} describes the datasets and methodologies used in our benchmarking experiments as well as the evaluation strategy and profiling method. In Section \ref{sec:results}, we present our experimental results. Finally, Section \ref{sec:discussion} discusses the applicative outcomes of these findings and suggests promising future research directions.

\section{Experimental setup}
\label{sec:experimental_setup}
In this section, we formally outline the experimental setup used to explore when DL models outperform ML models for tabular data. Initially, we present the datasets included in the analysis. Afterwards, the ML and DL models considered for the analysis are introduced, followed by the evaluation strategy for both the classification and regression tasks. Finally, the proposed profiling and meta-analysis analysis of which datasets DL models produce comparably better results than ML models. 

\subsection{Datasets}
In order to create a diverse and comprehensive benchmark for tabular datasets, aiming to provide insights into the scenarios where DL models outperform ML models, we incorporated a wide array of datasets exhibiting considerable variability and diversity of real-world tasks and characteristics. The benchmark includes datasets for both regression and classification tasks sourced from various domains such as economics and medicine to ensure relevance across different application areas. Additionally, we selected datasets of varying sizes in terms of both the number of rows (43-245,057) and columns (4-267), which are known to be crucial to the performance of ML and DL models, from previous studies. 

In our selection process, we also prioritized datasets containing categorical features. Categorical features are prevalent in real-world datasets and pose unique challenges, often necessitating specific preprocessing and modeling approaches. Ensuring their inclusion allows our benchmark to accurately reflect real-world complexities and facilitates an assessment of how well different models handle such features. Moreover, we ensured that our dataset selection varies in terms of difficulty, with some datasets presenting straightforward predictive tasks while others pose more complex challenges with respect to model prediction accuracy. This varying degree of difficulty ensures that our benchmark can comprehensively evaluate and differentiate model performance across simpler as well as more challenging predictive tasks. None of the datasets in this benchmark allowed a perfect prediction with all models. In Table \ref{tab:comparison_studies} we compare the characteristics of our datasets with previous works.

\begin{table}[ht]
\centering
\caption{Comparison with Previous Studies}
\label{tab:comparison_studies}
\small
\begin{tabular}{l@{\hskip 0.08in}c@{\hskip 0.08in}c@{\hskip 0.08in}c@{\hskip 0.08in}c@{\hskip 0.08in}c@{\hskip 0.08in}c@{\hskip 0.08in}c}
\toprule
\textbf{Study} & \textbf{\# Models} & \textbf{\#Classification} & \textbf{\#Regression} & \textbf{Median Dataset} & \textbf{Median \#} & \textbf{Median \# of} & \textbf{Meta-} \\
 &  & \textbf{Datasets} & \textbf{Datasets} & \textbf{Size} & \textbf{Features} & \textbf{Categorical} & \textbf{Analysis} \\
\midrule
\cite{why_do} & 7 & 22 & 33 & 17k & 13 & 1 & No \\
\cite{when_do} & 12-19* & 176 & 17 & 2k & 21 & 0 & Partial** \\
\cite{DL_not_all_you_need} & 6 & 9 & 2 & 11k & 32 & 1 & No \\
\cite{tabular_issue} & 19 & 4 & 1 & 20k & 21 & 2 & No \\
\cite{latest_review} & 5 & 181 & 119 & 12K & 21 & 1 & No \\
Ours & 20 & 54 & 57 & 5k & 13 & 4 & Yes \\
\bottomrule
\addlinespace
\multicolumn{8}{l}{\parbox[t]{\textwidth}{*12 models for regression datasets, 19 models for classification datasets.\\ **\cite{when_do} examine the effect of several meta-features, but do not present a predictive model for DL advantage over ML.}}
\end{tabular}
\end{table}

Overall, we included 111 datasets in this study: 57 regression datasets and 54 classification datasets. Table \ref{tab:descriptive_stats} summarizes the main parameters of the datasets. The full table of datasets is presented in the appendix. All datasets can be freely accessed online: 84 (76\%) of the datasets are obtained from OpenML; an additional 20 (18\%) regression datasets were obtained from a Materials Science regression benchmark \cite{reg_datasets}, and 7 (6\%) additional datasets were obtained from Kaggle.

\begin{table}[ht]
\centering
\caption{Descriptive Statistics of the Dataset Variables}
\hspace*{-0.08\textwidth}
\label{tab:descriptive_stats}
\begin{tabular}{lrrrrrrr}
\toprule
\textbf{Variable} & \textbf{Mean} & \textbf{STD} & \textbf{Min} & \textbf{25\%} & \textbf{Median} & \textbf{75\%} & \textbf{Max} \\
\midrule
Number of Rows & 18576 & 39874 & 43 & 673.75 & 4720 & 14057.5 & 245057 \\
Number of Columns & 24.16 & 40.53 & 4 & 8.75 & 12.5 & 22.25 & 267 \\
Number of Numerical Columns & 14.25 & 30.21 & 0 & 3 & 7 & 17 & 247 \\
Number of Categorical Columns & 9.91 & 24.76 & 0 & 1 & 4 & 9 & 231 \\
Kurtosis & 348.01 & 1129.66 & -2711.83 & -0.31 & 6.44 & 684.99 & 8901.75 \\
Average Correlation Between Features & 0.08 & 0.12 & -0.16 & 0.01 & 0.06 & 0.14 & 0.62 \\
Average Entropy & 7.70 & 2.29 & 2.45 & 6.07 & 7.96 & 9.33 & 14.17 \\
Average Pearson to Target Feature & 0.11 & 0.10 & -0.19 & 0.04 & 0.09 & 0.18 & 0.44 \\
\bottomrule
\end{tabular}
\end{table}
\subsection{Machine learning and deep learning models}
\label{sec:algos}
We use 20 ML and DL models to capture a representative set of algorithms from both groups for our analysis. These models are mainly adopted from previous benchmarking studies and due to their overall popularity \cite{when_do,why_do}. Formally, for tree-based ensemble models, we use XGBoost \cite{xgboost}, Random Forest \cite{rf}, AdaBoost \cite{adaboost}, LightGBM \cite{lightgbm}, CatBoost \cite{catboost}, and the H2O-GBM AutoML library \cite{h2o}. For DL models we use two AutoDL libraries (AutoGluon \cite{autogluon} and H2O \cite{h2o}, both restricted to DL models), a ResNet-like model \cite{resnet}, MLP \cite{mlp}. Additional models include AutoML libraries (TPOT \cite{tpot}, AutoGluon without the DL restriction \cite{autogluon}), and also SVM \cite{svm_fs}, KNN \cite{knn}, Decision Tree \cite{dt}, a symbolic regression model (GPLearn) \cite{tanemura2022application}, and Linear Regression or Logistic Regression \cite{lr_model} for regressions and classifications, respectively. All details of the models' runs are detailed in the appendix.

\subsection{Evaluation strategy}
Inspired by \cite{when_do}, we present the mean results of a 10-fold cross-validation evaluation. For the regression tasks we calcualte root mean squared error (RMSE), mean absolute error (MAE), and coefficient of determination ($R^2$). For the classification tasks, we present accuracy, Area Under the receiver operating characteristic (ROC) Curve (AUC), and \(F_1\) score. For each dataset, we evaluate the performance of each one of the models. We then present the performance summaries and ranking of each model. We also calculate the ranking by model groups (GBM/ML, DL, and others).

\subsection{Meta-analysis profiling}
In order to analyze for which datasets DL models outperform ML models, we adopted a meta-learning approach, following the same analysis proposed by \cite{teddy_meta_analysis_base}. Namely, we aim to find a meta-learning ML algorithm \((A^*)\) that receives as input a set of datasets \((D)\), a set of ML models \( (ML)\), and a set of DL models \((DL)\). It outputs a model (e.g., function) \((M)\) such that given a new dataset and the same sets of ML and DL models, the model \((M)\) returns whether ML or DL is best performing on the given dataset, according to some loss function \((L)\). Formally, the algorithm \(A^*\) satisfies:
\begin{equation}
    A^* := \min_{A \in \mathbb{A}} \Sigma_{d \in D} L \big ( A(d, ML, DL) \big ),
    \label{eq:optimal_ml}
\end{equation}
where \(\mathbb{A}\) is the set of all possible meta-learning models and \(A \in \mathbb{A}\) is a meta-learning model. We solve this optimization problem using a meta-learning approach. First, we construct a meta-dataset which operates as the data for the learning model. Second, we find a learning model that optimizes Eq. (\ref{eq:optimal_ml}) using a search algorithm. 

In order to obtain \(A^*\), we propose a meta-learning approach that requires a dataset to learn from. Thus, we constructed a meta-dataset as follows. First, each dataset is converted into a meta-feature vector with 20 parameters (full description provided in Table \ref{table:dataset_features}), marked as \(\Bar{X}\). This feature space is constructed from a simple feature \cite{feature_vector_1} such as the number of records and features, statistical properties of the dataset itself \cite{feature_vector_2} such as the fourth standardized moment, and statistical features measuring the connections between the independent features and the target feature \cite{feature_vector_3} such as the average Pearson correlation between the independent features and the target feature. These features have been used to obtain good results in previous meta-learning tasks \cite{feature_vector_1,feature_vector_2,feature_vector_3}. In addition, we compute the performance of each of the 20 algorithms (see Section \ref{sec:algos}) on the dataset (RMSE for regression, and AUC for classification), taking the model with the best performance. If this model belongs to the ML algorithms group, the target column (\(\Bar{Y}\)) of the meta-learning is set to 1 and 0 otherwise. Based on these two sets (\(\Bar{X}, \Bar{Y}\)), we define a meta-dataset such that (\(\Bar{X}\)) are the source features and (\(\Bar{Y}\)) is the target feature of the dataset.

Classifying whether ML or DL will best perform for a given dataset is a binary classification problem. We formalize this task as a search problem in which one needs to find the optimal configuration, as defined in the ML pipeline in Eq. (\ref{eq:optimal_ml}). One way to solve this classification problem is by using machine learning models. To this end, we used both a symbolic regression model and a ML model and  to obtain both explainability and investigate the best possible theoretical prediction, respectively. We used aggressive grid search optimization for the hyperparameters of both models, as well as the 10-fold cross-validation to ensure robustness. Specifically, due to the characteristics of this meta-dataset, we chose H2O-DL as the main model for this task.

\section{Results}
\label{sec:results}
In this section, we present the results of the benchmarking analysis. We begin by outlining the performance of each of the models on the datasets, summarizing 4,000 computation hours. Afterward, we explore the influence of several central properties of datasets and their influence on ML and DL performance. Finally, we provide a measurable profiling of when DL outperforms ML models on tabular tasks.

\subsection{Model Ranking}
\label{sec:ranking}
Table \ref{tab:ranking_tedl} outlines the performance of the Tree-based Ensemble models (TE) and DL models on the 111 datasets. The models are ordered from top to bottom according to the number of datasets where they outperformed the other datasets. One can notice that ML models occupy the first four lines, led by CatBoost with 19/111 (17.1\%). The first DL model appears on the fifth row with 11/111 (9.9\%) datasets where it is best-performing. This ranking is preserved by other metrics such as average ranking and median ranking.

\begin{table}[h!]
\centering
\caption{Performance ranking of TE and DL models.}
\begin{tabular}{lccccc}
\hline\hline
Model & Group & \# Best & Average Rank & Median Rank & \# in Top 3 Models \\
\hline\hline
CatBoost & TE & 19 & 4.9 & 4 & 50 \\
LightGBM & TE & 15 & 5 & 4 & 47 \\
H2O-GBM & TE & 13 & 7 & 6 & 28 \\
Random Forest & TE & 11 & 6.7 & 6.5 & 28 \\
AutoGluon-DL & DL & 11 & 6.6 & 7 & 32 \\
ResNet & DL & 10 & 7.5 & 8 & 23 \\
TPOT & TE & 7 & 5.9 & 5 & 39 \\
H2O-DL & DL & 6 & 8.8 & 9 & 13 \\
MLP & DL & 6 & 7.9 & 8 & 17 \\
XGBoost & TE & 5 & 6.4 & 6 & 34 \\
AdaBoost & TE & 4 & 10.1 & 11 & 9 \\
DCNV2 & DL & 4 & 9 & 10 & 12 \\
FT-Transformer & DL & 0 & 10.7 & 11 & 1 \\
TabNet & DL & 0 & 13.1 & 14 & 0 \\
\hline\hline
\end{tabular}
\label{tab:ranking_tedl}
\end{table}

\begin{figure}[!ht]
    \centering
\includegraphics[width=0.92\textwidth]{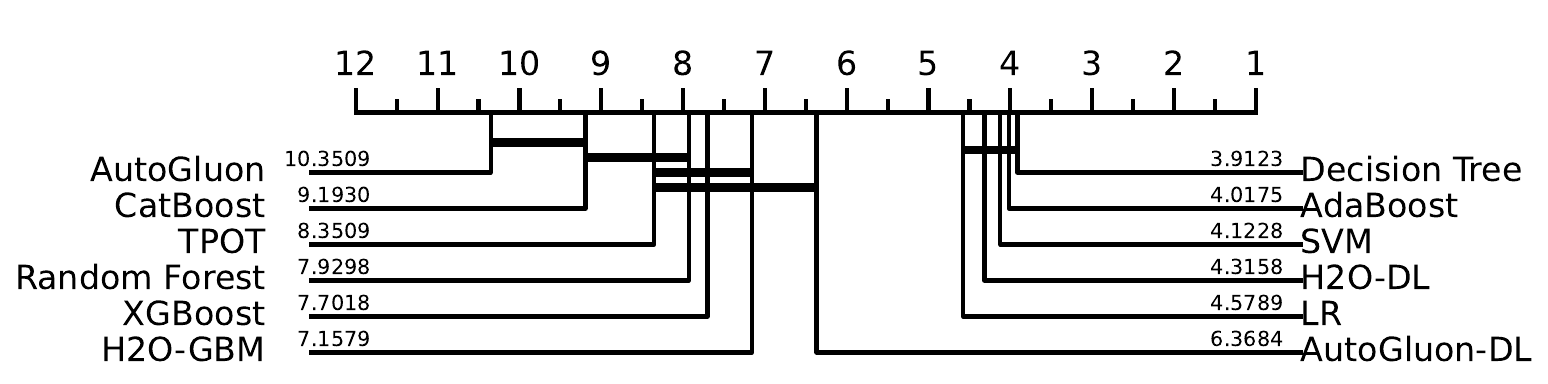}
    \caption{Critical difference diagram for regression tasks based on RMSE. The best performing model is AutoGluon as lower RMSE scores indicate better performance.}
    \label{fig:cd_rmse}
\end{figure}

\begin{figure}[!ht]
    \centering
\includegraphics[width=0.92\textwidth]{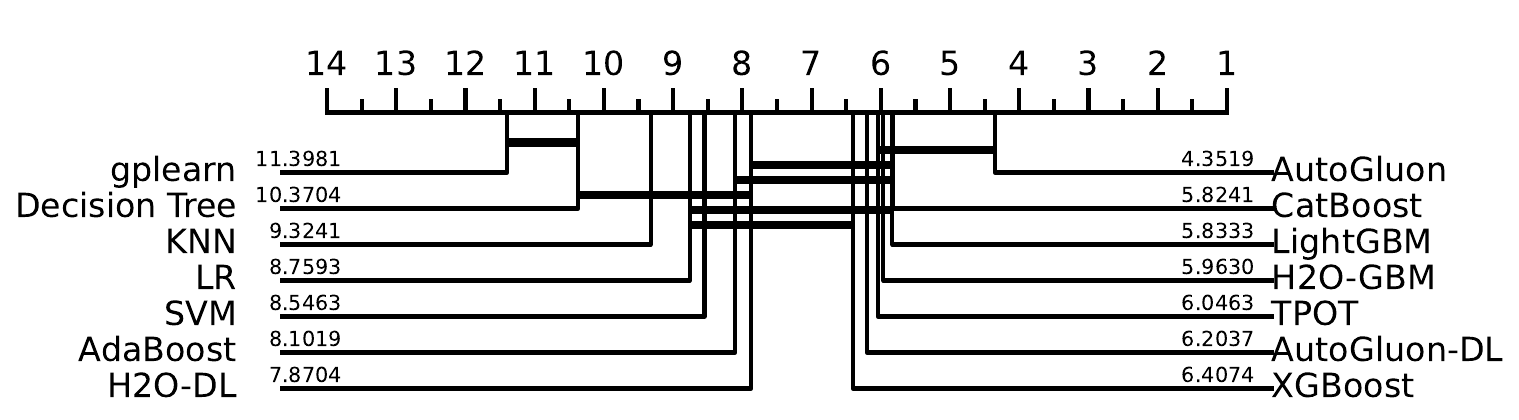}
    \caption{Critical difference diagram for classification tasks based on accuracy. The best performing model is AutoGluon as higher accuracy scores indicate better performance.}
    \label{fig:cd_acc}
\end{figure}

Table \ref{tab:ranking_all} extends Table \ref{tab:ranking_tedl} as it presents the performance of all 20 models on the 111 datasets. Notably, AutoGluon, an ensemble of DL and ML models method best performed for 39 out of the 111 datasets (35\%), almost four times more than SVM, the second-best model, which best-performed in only 10 datasets (9\%). To this end, AutoGluon is also the best-performing model, on average, while SVM's is actually poorly performing on average and excels occupancy. In a more general sense, ML models occupy the top three most performing models of each dataset most of the time (see last column). TabNet is reaching last place with no datasets where it is the best-performing one with also the worst average position from all the models. FT-Transformer shows similar behavior, only slightly outperforming TabNet. The results are also summarized using critical difference diagrams based on RMSE for regressions tasks (Fig. \ref{fig:cd_rmse}) and on accuracy for classifications tasks (Fig. \ref{fig:cd_acc}). The performance ranking based on alternative metrics (MAE, $R^2$, AUC, and F1-score) are presented in the appendix.

\begin{table}[h!]
\centering
\caption{Performance ranking of all models.}
\begin{tabular}{lcccccc}
\hline\hline
Model & Group & \# Best & Average Rank & Median Rank & \# in Top 3 Models \\
\hline\hline
AutoGluon & Other & 39 & 4.8 & 4 & 58 \\
SVM & Other & 10 & 12.4 & 14 & 15 \\
ResNet & DL & 7 & 9.7 & 10 & 13 \\
CatBoost & TE & 7 & 6.6 & 5 & 35 \\
LightGBM & TE & 6 & 6.9 & 6 & 33 \\
H2O-GBM & TE & 6 & 8.6 & 8 & 18 \\
TPOT & TE & 5 & 7.7 & 7 & 23 \\
AutoGluon-DL & DL & 5 & 8.7 & 8 & 21 \\
H2O-DL & DL & 4 & 11.5 & 11 & 11 \\
gplearn & Other & 3 & 15 & 17 & 7 \\
MLP & DL & 3 & 9.6 & 10 & 13 \\
LR & Other & 3 & 11.6 & 13 & 16 \\
XGBoost & TE & 3 & 8.4 & 8 & 19 \\
Random Forest & TE & 3 & 8.5 & 8 & 20 \\
DCNV2 & DL & 3 & 11.6 & 12 & 10 \\
KNN & Other & 2 & 12.1 & 13 & 12 \\
Decision Tree & Other & 1 & 13.3 & 14 & 3 \\
AdaBoost & TE & 1 & 12.3 & 13 & 5 \\
FT-Transformer & DL & 0 & 13.9 & 14 & 1 \\
TabNet & DL & 0 & 17.2 & 18 & 0 \\
\hline\hline
\end{tabular}
\label{tab:ranking_all}
\end{table}

Table \ref{tab:small_tedl} follows the same line but includes only datasets with less than 1000 rows (36). In this scenario, H2O led the chart with 6/36 (16.6\%), followed closely by ResNet with 5/36 (13.9\%) datasets in which they are best performing. However, ResNet lags behind other ML models in the other metrics (average rank, median rank, and top 3 models) compared to CatBoost in the third row. Interestingly, ResNet shows similar performance to the much simpler MLP model as well as the more complex AutoGluon-DL model. 

\begin{table}[h!]
\centering
\caption{Performance metrics of TE and DL models for small datasets ($<1000$) rows.}
\begin{tabular}{lccccc}
\hline\hline
Model & Group & \# Best & Average Rank & Median Rank & \# in Top 3 Models \\
\hline\hline
H2O-GBM & TE & 6 & 5.6 & 5 & 12 \\
ResNet & DL & 5 & 6.7 & 7 & 8 \\
CatBoost & TE & 4 & 5.1 & 5 & 14 \\
AutoGluon-DL & DL & 4 & 6.8 & 7.5 & 10 \\
MLP & DL & 4 & 6.3 & 6 & 9 \\
TPOT & TE & 3 & 5.1 & 4 & 15 \\
Random Forest & TE & 3 & 6.1 & 5.5 & 7 \\
LightGBM & TE & 3 & 4.9 & 4 & 13 \\
XGBoost & TE & 2 & 6.6 & 7.5 & 13 \\
H2O-DL & DL & 1 & 9.6 & 11 & 2 \\
DCNV2 & DL & 1 & 9.1 & 10 & 2 \\
AdaBoost & TE & 0 & 8.9 & 8.5 & 2 \\
FT-Transformer & DL & 0 & 10.1 & 11 & 1 \\
TabNet & DL & 0 & 13.9 & 14 & 0 \\
\hline\hline
\end{tabular}
\label{tab:small_tedl}
\end{table}

\subsection{Meta-Analysis Profiling}
\label{sec:feature_importance}

We now present the results of the prediction of whether ML or DL will perform better in each dataset. Table \ref{table:logistic_regression} presents the coefficients of a logistic regression for this classification task. The model reveals several important findings. First, it demonstrates with statistical significance that the relative performance of DL models (compared to TE models) is better in classification tasks than in regression tasks. Second, we find that the Kurtosis variable is statistically significant. Finally, we find that the PCA components are also positive and almost statistically significant. We discuss these findings in Section \ref{sec:discussion}.

Next, we repeat this analysis after limiting the datasets to cases where the performance of ML/DL models was significantly different (with $p<0.05$). This restriction leaves 36 datasets, 11 of which demonstrate higher performance for DL and 25 for TE. As we show later on, and as previously noted by \cite{when_do}, DL models have an advantage in small datasets; we therefore use the H2O-DL model to predict whether TE or DL performs better based on the properties of each dataset. Remarkably, H2O-DL provides a high performance in this classification task, with an AUC of 0.78, accuracy of 86.1\%, and F1 score of 0.61. As a baseline, we also train a logistic regression, an explainable model, which obtains lower but still impressive performance (AUC of 0.68, accuracy of 80.6\%, and F1 score of 0.44).

\begin{table}[!ht]
\centering
\caption{Coefficients of a logistic regression for predicting the probability that DL outperforms ML.}
\begin{tabular}{lcccc}
\hline \hline
\textbf{Variable} & \textbf{Coefficient} & \textbf{Std Error} & \textbf{z-value} & \textbf{P>|z|} \\ 
\hline \hline
Intercept & -0.8751 & 0.2345 & -3.7315 & 0.0002 \\ 
Row count & -0.0195 & 0.5688 & -0.0342 & 0.9727 \\ 
Row over Column & -1.5991 & 1.6984 & -0.9415 & 0.3464 \\ 
Classification/Regression & 0.5563 & 0.2590 & 2.1483 & 0.0317 \\ 
Cancor & -0.2067 & 0.2584 & -0.8000 & 0.4237 \\ 
Kurtosis & 0.8975 & 0.3987 & 2.2514 & 0.0244 \\ 
Average Asymmetry Of Features & -0.0316 & 0.2480 & -0.1274 & 0.8986 \\ 
Average Pearson to Target Feature & 0.5247 & 0.2984 & 1.7581 & 0.0787 \\ 
Standard Deviation Pearson To Target Feature & -0.2326 & 0.2772 & -0.8390 & 0.4014 \\ 
Average Correlation Between Features & -0.1639 & 0.2501 & -0.6552 & 0.5123 \\ 
Average Coefficient of Variation & -0.1087 & 0.4039 & -0.2692 & 0.7878 \\ 
Standard Deviation Coefficient of Variation & -0.0320 & 0.4196 & -0.0764 & 0.9391 \\ 
Average Coefficient of Anomaly & 0.0588 & 0.2746 & 0.2140 & 0.8305 \\ 
Standard Deviation Coefficient of Anomaly & -0.2730 & 0.2859 & -0.9550 & 0.3396 \\ 
Average Entropy & -0.2086 & 0.2708 & -0.7704 & 0.4410 \\ 
Standard Deviation Entropy & 0.1769 & 0.2572 & 0.6877 & 0.4916 \\ 
Columns after One Hot Encoding & -0.5745 & 0.3948 & -1.4550 & 0.1457 \\ 
Rows over Columns after One Hot Encoding & 1.7303 & 1.4779 & 1.1708 & 0.2417 \\ 
PCA & 0.4624 & 0.2648 & 1.7462 & 0.0808 \\ 
\hline \hline
\end{tabular}
\label{table:logistic_regression}
\end{table}

\begin{figure}[!ht]
    \centering
\includegraphics[width=0.92\textwidth]{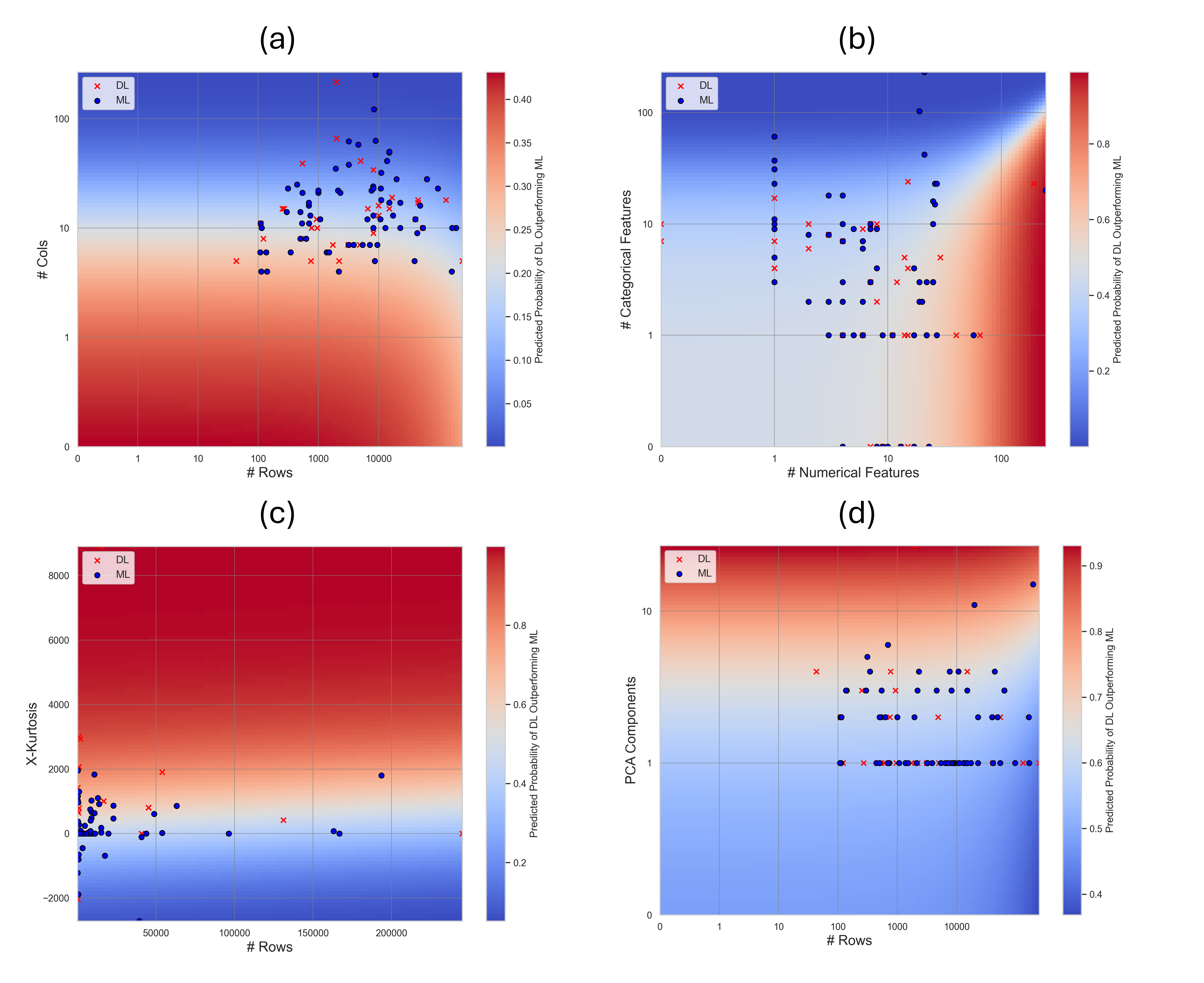}
    \caption{The effect of various factors on the probability that DL outperforms ML. The heatmaps are generated using the prediction of the logistic regression models. The scatter plot represents the actual observations of the datasets. (a) the impact of the number of columns and rows; (b) the influence of numerical and categorical feature counts; (c) the effect of X-kurtosis and row count; and (d) the role of PCA components necessary to maintain 99\% of the variance and number of rows.}
    \label{fig:heatmaps}
\end{figure}

Based on the predictions of the logistic regression model, we provide further insights into the most influential factors for TE versus DL performance. Fig. \ref{fig:heatmaps} presents heatmaps of four dataset's configurations and their influence on the probability that a DL model would outperform the ML model for a given dataset, including (a) the impact of the number of columns and rows; (b) the influence of numerical and categorical feature counts; (c) the effect of X-kurtosis and row count; and (d) the role of PCA components necessary to maintain 99\% of the variance. As one can see from sub-figure (a), for a small number of rows, increasing the number of columns results in a higher probability that the DL model would outperform an ML model. However, this effect decreases relatively quickly as the number of rows increases. Notably, for all the explored configurations, the probability does not increase over 0.5 which indicates no configuration found where DL models would outperform ML models, on average. For sub-figure (b), a more clear gradient is revealed where a smaller number of rows and a higher number of columns increase the probability that DL models outperform ML models. Interestingly, this heatmap reveals configurations where the probability is higher than 0.5. For sub-figure (c), one can notice that the number of rows does have much influence on the probability while a large X-kurtosis signals that DL models are probably preferred over ML models. Finally, sub-figure (d), shows a similar trend to the previous sub-figures where a small number of rows and columns, especially if these are more \say{computationally-attractive} like PCA-based features, results in a higher probability for DL models to outperform ML models. 

Finally, we trained a symbolic regression (SR) model to search for more complex equations of this prediction task. The performance of the SR model was relatively close to that of the logistic regression. The formula output, which we present in Eq. \ref{eq:logreg}, also predicts higher relative DL success rate in small datasets and with high Kurtosis values.

\begin{equation}
\label{eq:logreg}
\text{logreg} \left( 0.005 \cdot x_{\text{kurtosis}} - 4.3 \times 10^{-5} \cdot x_{\text{row\_count}} - 0.053 \cdot x_{\text{std\_coefficient\_of\_anomaly}} - 23.0 \cdot x_{\text{std\_linearly\_to\_target}} + 0.89 \right)
\end{equation}

where,

\begin{align*}
\text{logreg}(z) &= \frac{1}{1 + e^{-z}} \\
x_{\text{kurtosis}} &\text{ is the kurtosis} \\
x_{\text{row\_count}} &\text{ is the number of rows} \\
x_{\text{std\_coefficient\_of\_anomaly}} &\text{ is the standard deviation of the coefficient of anomaly} \\
x_{\text{std\_linearly\_to\_target}} &\text{ is the standard deviation linearly to the target}
\end{align*}

Next, to explore the effect of data size, we repeated the training of 10 large datasets after sampling them to 1000 training samples. The original and revised rankings for these 10 datasets are summarized in Table \ref{table:comparison}. While the ranking of some DL models (AutoGluon-DL and ResNet) improves in the smaller datasets, TE models still dominate this set. Although this is a relatively small sample of 10 datasets, it provides further evidence that sample size is an important factor but does not determine which model will have the best performance by itself.

\begin{table}[]
\caption{Comparison of model performance on original and sampled datasets}
\label{table:comparison}
\small
\centering
\setlength{\tabcolsep}{3pt}
\begin{tabular}{lcccccc}
\toprule
\multicolumn{6}{c}{\textbf{Original}} \\
\cmidrule(lr){1-6}
\textbf{Model} & \textbf{Group} & \textbf{\# Best} & \textbf{Average Rank} & \textbf{Median Rank} & \textbf{\# in Top 3 Models} \\
\midrule
LightGBM & TE & 3 & 4.1 & 3 & 5 \\
CatBoost & TE & 2 & 3.3 & 3.5 & 5 \\
XGBoost & TE & 1 & 4.5 & 4 & 3 \\
Random Forest & TE & 1 & 4.3 & 3 & 6 \\
H2O-GBM & TE & 1 & 6 & 6 & 2 \\
H2O-DL & DL & 1 & 8 & 9 & 2 \\
MLP & DL & 1 & 8.1 & 8 & 1 \\
TPOT & TE & 0 & 5.7 & 4 & 4 \\
AdaBoost & TE & 0 & 13.2 & 13 & 0 \\
AutoGluon-DL & DL & 0 & 6.9 & 7 & 1 \\
ResNet & DL & 0 & 7.3 & 9 & 1 \\
FT-Transformer & DL & 0 & 10.8 & 11 & 0 \\
TabNet & DL & 0 & 13.8 & 14 & 0 \\
DCNV2 & DL & 0 & 14.5 & 14.5 & 0 \\
\midrule
\multicolumn{6}{c}{\textbf{Sampled to 1,000 Rows}} \\
\cmidrule(lr){1-6}
\textbf{Model} & \textbf{Group} & \textbf{\# Best} & \textbf{Average Rank} & \textbf{Median Rank} & \textbf{\# in Top 3 Models} \\
\midrule
LightGBM & TE & 4 & 2.4 & 2 & 7 \\
XGBoost & TE & 2 & 4.2 & 3 & 6 \\
TPOT & TE & 1 & 4.2 & 4 & 4 \\
CatBoost & TE & 1 & 3.6 & 3.5 & 5 \\
AutoGluon-DL & DL & 1 & 7.6 & 7.5 & 2 \\
ResNet & DL & 1 & 6.9 & 8.5 & 2 \\
Random Forest & TE & 0 & 6.5 & 6 & 1 \\
AdaBoost & TE & 0 & 11.7 & 12 & 0 \\
H2O-GBM & TE & 0 & 6.1 & 5.5 & 2 \\
H2O-DL & DL & 0 & 9.5 & 9 & 0 \\
MLP & DL & 0 & 7.6 & 8 & 1 \\
FT-Transformer & DL & 0 & 10.9 & 11 & 0 \\
TabNet & DL & 0 & 13.7 & 14 & 0 \\
DCNV2 & DL & 0 & 12.4 & 13.5 & 0 \\
\bottomrule
\end{tabular}
\end{table}

\section{Discussion}
\label{sec:discussion}
In this study, we benchmark the performance of 20 data-driven models, divided between ML and DL models, for the tasks of regression and classification of tabular data from 111 datasets. While DL models are currently state-of-the-art in multiple computational tasks such as computer vision, natural language processing, and signal processing, to name a few, they are outperformed by ML models for tabular data. Following several recent benchmark studies, we computed one of the most comprehensive benchmarking of ML and DL models' performance on tabular data and their profiling. In this study, we focused on providing measurable (statistical) features of datasets, available before running any model, which can indicate when DL models would outperform ML models in both regression and classification tasks. The trained model obtained a remarkable accuracy of 86.1\% and AUC of 0.76. We also presented explainable models, logistic regression and symbolic regression, with slightly lower performance.

Similar to previous benchmarking studies \cite{Benchmarking_AutoML,DL_not_all_you_need}, our analysis shows that ML models, on average, outperform DL models on tabular data. Specifically, Tree-based Ensemble models consistently exhibit the highest performance in this field. This behavior is consistent for the best-performing model, as well as the mean rank, and 3-top model, indicating that on a random tabular dataset, ML models would be the safe \say{bate}, as indicated by Table \ref{tab:ranking_all}. In particular, we found that AutoGluon, an automatic ML model \cite{substract} that uses ensembles of both ML and DL models, outperforms the other models by a large margin. This outcome aligns with the findings presented by \cite{when_do}. In addition, we found that TabNet actually performs worse than DL models that are not specifically designed for tabular data such as MLP and H2O, which agrees with previous attempts of using TabNet as part of benchmarking attempts \cite{simple_MLPs}. 

Moreover, previous studies do not show clear agreement on the influence of the number of rows and cols on the probability that DL models would outperform ML models \cite{why_do,DL_not_all_you_need,when_do}, we tackle this challenge as shown in Fig. \ref{fig:heatmaps}. Our results clearly indicate a somewhat linear trend where a smaller number of rows and a larger number of columns, on average, results in a higher probability that DL models would outperform ML models. However, the analysis in Table \ref{table:comparison} which tries to isolate the size factor does not show clear results; our conclusion is that while DL models may outperform other ML models in small datasets in many cases, this is just one of many other factors which affect the relative performance of the two model groups.

Regarding the profiling of the dataset characteristics to predict whether either DL model or ML model would provide the best-performing results, a logistic regression analysis (see Table \ref{table:logistic_regression}) reveals only two statistically significant dataset's features - is the task a regression or classification and the kurtosis metric. In particular, the relative performance of DL models compared to other ML models is better in classification tasks compared to regression tasks. This is also highlighted by the separate regression and classification ranking in the appendix (Tables \ref{tab:rmse_tedl}, \ref{tab:auc_tedl}). A possible explanation for this phenomenon is that in classification tasks, all errors contribute equally to the performance metric while in regression tasks, large errors have more weight (RMSE is unbounded). Due to the large parameter space of DL models, it is possible that they could sometimes exhibit large errors which are heavily penalized by metrics such as RMSE. Indeed, when ranking the models by MAE for regression tasks, the relative ranking of DL models improves significantly with AutoGluon-DL positioned second after CatBoost (Table \ref{tab:mae_tedl}). Finally, focusing on the kurtosis metric, a large value indicates a long \say{tail} distribution where DL is known to excel compared to ML models \cite{last_power_law}.

\textbf{Limitations and future work.} While this study presented an exhaustive evaluation of the different models over 111 datasets, it still has several limitations. First, the choice to include diverse datasets, in contrast to several previous benchmarks, has many advantages but also some disadvantages. For example, including small datasets could introduce more noise into the results. In addition, including many types of datasets inevitably means each type will have fewer instances. We tackled this problem by including a large number of datasets, but including one type of homogeneous datasets would obviously result in more instances for this type. Second, an analysis of feature selection or feature engineering, which is known to have a significant impact on the down-the-line model's performance, has not been included in this work. Finally, while this study included diverse regression and classification datasets, there are additional tasks that have not been included, such as time-series or multilabel classifications. These extensions can reveal additional insights regarding the performance of ML and DL on tabular data and are promising research future venues.  

\bibliography{biblio}
\bibliographystyle{unsrt}

\section{Supplementary Material}
\subsection{Description of models}
\label{models}

\textbf{TPOT} is an open-source library that automates the process of designing and optimizing ML pipelines. It uses genetic programming to explore a wide range of models and preprocessing steps, aiming to find the best pipeline for a given dataset. TPOT also performs hyperparameter optimization. We ran the AutoML library with mostly default settings, limiting each model's run to one hour.

\textbf{H2O} is an open-source software for data analysis that facilitates the development and deployment of machine learning models. It provides a scalable and fast platform for building models, with support for a variety of algorithms, including generalized linear models, gradient boosting, and deep learning. H2O's automated machine learning functionality assists in discovering the best models by automatically training and tuning multiple models within a user-specified time frame. We ran two H2O models, both the H2O Gradient Boosting Machine (labeled H2O-GBM) and the H2O Deep Learning (labeled H2O-DL). Both were limited to one hour for each model run.

\textbf{XGBoost} is an open-source ML algorithm developed for supervised learning tasks. It is an implementation of gradient boosted decision trees. Technically, XGBoost enhances performance by optimizing for speed and scalability, using techniques like parallel processing, tree pruning, and regularization to prevent overfitting. It also supports missing value handling and a range of objective functions. Widely recognized for its superior predictive power and fast execution, XGBoost has been a top choice in ML tabular tasks on a wide range of domains. We performed hyperparameter optimization using TPOT, with a one hour time limit. 

\textbf{Random Forest} extends the bagging method by incorporating both bagging and feature randomness to generate an uncorrelated ensemble of decision tree models. Feature randomness creates a random subset of features, ensuring low correlation among the decision trees which improves the generalization of Random Forest compared to other bagging decision tree models. We performed hyperparameter optimization using TPOT, with a one hour time limit. 

\textbf{AdaBoost} classifier is a meta-estimator that starts by fitting a classifier, usually a decision tree, to the original dataset. It then fits additional copies of the classifier to the same dataset, adjusting the weights of incorrectly classified instances so that subsequent classifiers focus more on the challenging cases - a method usually called boosting. We performed hyperparameter optimization using TPOT, with a one hour time limit.

\textbf{CatBoost} short for Categorical Boosting, is an open-source boosting library. It is tailored for regression and classification problems with a large number of independent features. Unlike traditional gradient boosting methods, CatBoost can directly handle both categorical and numerical features without needing feature encoding techniques (like One-Hot Encoding or Label Encoding). Moreover, it employs an algorithm called symmetric weighted quantile sketch to automatically handle missing values, thereby reducing overfitting and enhancing the overall performance of the model. We performed hyperparameter optimization using TPOT, with a one hour time limit.

\textbf{Decision Tree} is a family of models that defines the connections between features and their possible consequences as a tree-like structure of nodes, where each internal node represents a feature (or attribute) test, each branch represents the outcome of the test, and each leaf node represents a class label (for classification) or a continuous value (for regression). The algorithm splits the dataset into subsets based on the feature that results in the highest information gain or the lowest impurity, according to the popular Scikit-learn library, but not limited to these options. This process occurs recursively until no further division is possible. We performed hyperparameter optimization using TPOT, with a one hour time limit.

\textbf{Linear Regression and Logistic Regression.} model the relationship between a set of features by fitting a linear equation to the observed data. Linear regression aims to minimize the sum of squared differences between the observed and predicted values. It is widely used due to its simplicity, interpretability, and efficiency, making it suitable for a variety of applications such as trend analysis, forecasting, and inferential statistics.

\textbf{GPLearn}, or Genetic Programming for scikit-learn, is an open-source Python library that applies genetic programming to ML tasks. It allows for the automatic generation of analytical (formula-based) models by evolving programs to fit data, using principles inspired by biological evolution such as selection, mutation, and crossover (also known as symbolic regression). GPlearn can be used for regression and classification problems, where it evolves mathematical expressions to optimize predictive performance. We configured the genetic programming algorithm with 50 generations and a parsimony coefficient of 0.001 to control model complexity.

\textbf{SVM} is a supervised ML algorithm used for classification and regression tasks. SVM works by finding the optimal hyperplane that best separates the data points of different classes in a high-dimensional space. This hyperplane maximizes the margin, which is the distance between the closest data points (support vectors) of each class. SVM can handle both linear and non-linear classification by using kernel functions to transform the input data into a higher-dimensional space where a linear separator can be found. Popular kernels include linear, polynomial, and radial basis functions. We performed hyperparameter optimization using TPOT, with a one hour time limit. 

\textbf{KNN} is a simple yet powerful supervised ML algorithm used for classification and regression tasks. It operates on the principle of similarity, where new instances are classified based on the majority class or average of the k nearest neighbors in the feature space. The algorithm does not involve explicit training; instead, it stores the entire training dataset and performs computations at prediction time. We performed hyperparameter optimization using TPOT, with a one hour time limit. 

\textbf{FT-Transformer} The FT-Transformer (FTT) model is a novel deep learning architecture specifically designed to handle tabular data effectively. It leverages the principles of the Transformer model, which is widely used in natural language processing, to capture the intricate relationships within tabular datasets. The model employs feature tokenization to handle numerical and categorical features and uses self-attention mechanisms to learn complex interactions between features. In our implementation, we used Optuna to optimize the hyperparameters of the FT-Transformer. As in \cite{issue_1}, we set the number of heads to 8. We varied the number of Transformer blocks (n\_blocks) between 1 and 6. The dimension of the token embeddings (d\_token) was set to be a multiple of 8, ranging from 8 to 32 times 8. The dropout rate for the attention mechanism (attention\_dropout) and the feed-forward network (ffn\_dropout) was varied between 0.1 and 0.5. The hidden dimension in the feed-forward network (ffn\_d\_hidden) was set between 64 and 256, and the residual dropout (residual\_dropout) was also varied between 0.1 and 0.5. The learning rate (learning\_rate) was set between 10\^-4 and 10\^-2, and the batch size (batch\_size) was chosen as one of 32, 64, or 128.

\textbf{ResNet} is a DL architecture designed to address the vanishing gradient problem in very deep neural networks. It introduces skip connections, also known as residual connections, that allow gradients to flow more effectively during training. ResNet architectures typically stack multiple residual blocks to form a deep network. The skip connections in each block enable the gradient to propagate more efficiently through the network, alleviating the vanishing gradient problem and enabling the training of very deep models. we utilized Optuna to optimize the hyperparameters by suggesting a range of values for each parameter. Specifically, we set the number of blocks (n\_blocks) between 1 and 5. For the main dimension of each block (d\_main), the value was set between 16 and 64, and for the hidden dimension (d\_hidden) within each block, the value was set between 16 and 64. The dropout rates for the first dropout layer (dropout\_first) and the second dropout layer (dropout\_second) were varied between 0.1 and 0.5. The learning rate (learning\_rate) was set between 10\^-4 and 10\^-2, and the batch size (batch\_size) was chosen as one of 32, 64, or 128. 

\textbf{MLP}, a Multilayer Perceptron is a type of artificial neural network that consists of multiple layers of interconnected nodes (neurons). MLPs are widely used for supervised learning tasks such as classification and regression. The network typically consists of an input layer, one or more hidden layers, and an output layer. Each neuron in an MLP performs a weighted sum of its inputs, followed by the application of an activation function to produce an output. The weighted sum is calculated by multiplying each input by a corresponding weight and summing them up, usually with the addition of a bias term. The activation function introduces non-linearity into the network, enabling it to learn complex relationships in the data. To train the MLP model in this code, we implemented a MLP model using PyTorch, and optimized its hyperparameters with Optuna to minimize the mean squared error on the validation set. We set the number of layers (n\_layers) between 1 and 5. For each layer, the number of units (n\_units\_l{i}) was set between 4 and 128, and dropout rates (dropout\_l{i}) for each layer were set between 0.1 and 0.5. The learning rate (learning\_rate) was varied between 10\^-4 and 10\^-2, and the batch size (batch\_size) was chosen as one of 32, 64, or 128.

\textbf{AutoGluon} is an open-source library for automated machine learning. AutoGluon supports a wide range of ML tasks, including classification, regression, and even object detection for computer vision applications. It is built on top of the deep learning framework Apache MXNet, providing scalability and efficiency for training models on large datasets. It automatically handles tasks such as feature selection, algorithm selection, hyperparameter tuning, and model ensembling. We ran either the \say{full} library (labeled as the AutoGluon model) or by restricting it to DL architectures (labeled as AutoGluon-DL) with a limit of 200 epochs.

\subsection{Meta-learning features}
Table \ref{table:dataset_features} mostly adopted from \cite{teddy_meta_analysis_base} and contains 20 features computationally useful for meta-learning tasks.

\begin{table}[!ht]
\centering
\caption{The meta-feature vector representing a dataset.}
\begin{tabular}{p{0.3\textwidth}p{0.6\textwidth}p{0.1\textwidth}}
\hline \hline
\textbf{Name} & \textbf{Description} & \textbf{Source} \\ \hline \hline
Row count & The number of records (rows) in the dataset. & \cite{feature_vector_1}\\ 
Column count & The number of features (columns) in the dataset. & \cite{feature_vector_1} \\ 
Columns after One Hot Encoding & The number of columns after one-hot encoding categorical features. & \cite{feature_vector_1} \\ 
Numerical Features & The number of numerical features in the dataset. & \cite{feature_vector_1} \\ 
Categorical Features & The number of categorical features in the dataset. & \cite{feature_vector_1} \\ 
Classification/Regression & The type of task, whether classification or regression. & \cite{feature_vector_2} \\ 
Cancor & Canonical correlation for the best single combination of features. & \cite{feature_vector_2} \\ 
Kurtosis & The fourth standardized moment. & \cite{feature_vector_2} \\ 
Average Entropy & The average entropy of the features in the dataset. & \cite{feature_vector_3} \\ 
Standard Deviation Entropy & The standard deviation entropy of the features in the dataset. & \cite{feature_vector_3} \\ 
Row Over Column & The number of records divided by the number of features in the dataset. & \cite{feature_vector_4} \\ 
Average Asymmetry Of Features & The average value of Pearson’s asymmetry coefficient. & \cite{feature_vector_3} \\ 
Average Pearson to Target Feature & The average Pearson correlation score of all the features in the dataset and the target feature. & \cite{feature_vector_3} \\ 
Standard Deviation Pearson To Target Feature & The standard deviation of the Pearson correlation scores between all the features in the dataset and the target feature. & \cite{feature_vector_3} \\ 
Average Correlation Between Features & The average Pearson correlation score between all the features. & \cite{feature_vector_3} \\ 
Average Coefficient of Variation & The average value of the standard deviation divided by the mean of each feature for all the features in the dataset. & \cite{feature_vector_3} \\ 
Standard Deviation Coefficient of Variation & The standard deviation value of the standard deviation divided by the mean of each feature for all the features in the dataset. & \cite{feature_vector_3} \\ 
Average Coefficient of Anomaly & The average value of the mean divided by the standard deviation of each feature for all the features in the dataset. & \cite{feature_vector_3} \\ 
Standard Deviation Coefficient of Anomaly & The standard deviation value of the mean divided by the standard deviation of each feature for all the features in the dataset. & \cite{feature_vector_3} \\
PCA & The number of PCA components required to explain 99\% of the variance in the data. & \cite{pca_1} \\\hline \hline
\end{tabular}
\label{table:dataset_features}
\end{table}

\subsection{Additional results}
In this section, we provide several additional results to further support the claims presented in this study. In particular, we provide the model's performance in several contexts such as larger datasets, few dimensions, and others.

\begin{figure}[!ht]
    \centering
\includegraphics[width=0.92\textwidth]{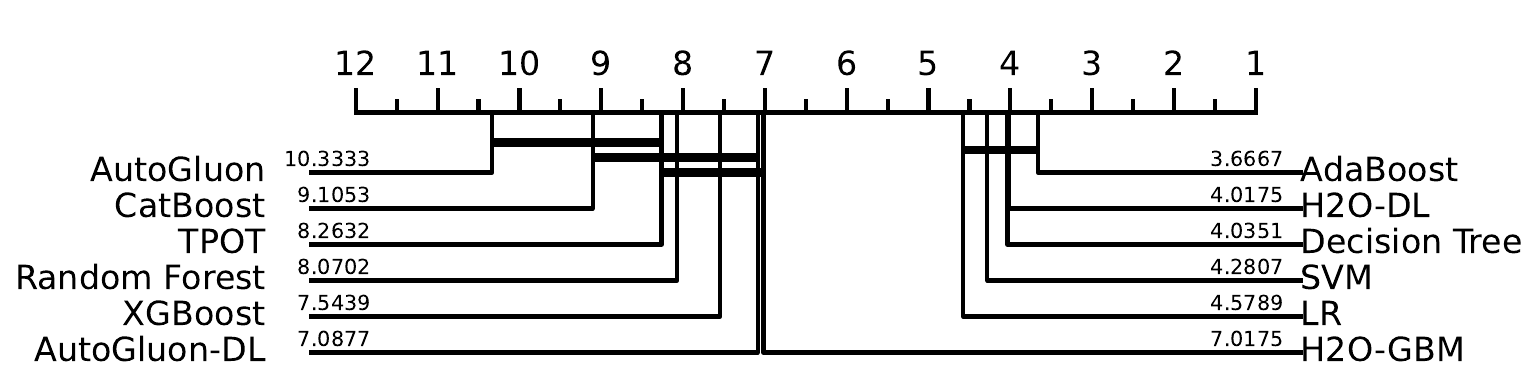}
    \caption{Critical difference diagram for regression tasks based on MAE. The best performing model is AutoGluon as lower MAE scores indicate better performance.}
    \label{fig:cd_mae}
\end{figure}

\begin{figure}[!ht]
    \centering
\includegraphics[width=0.92\textwidth]{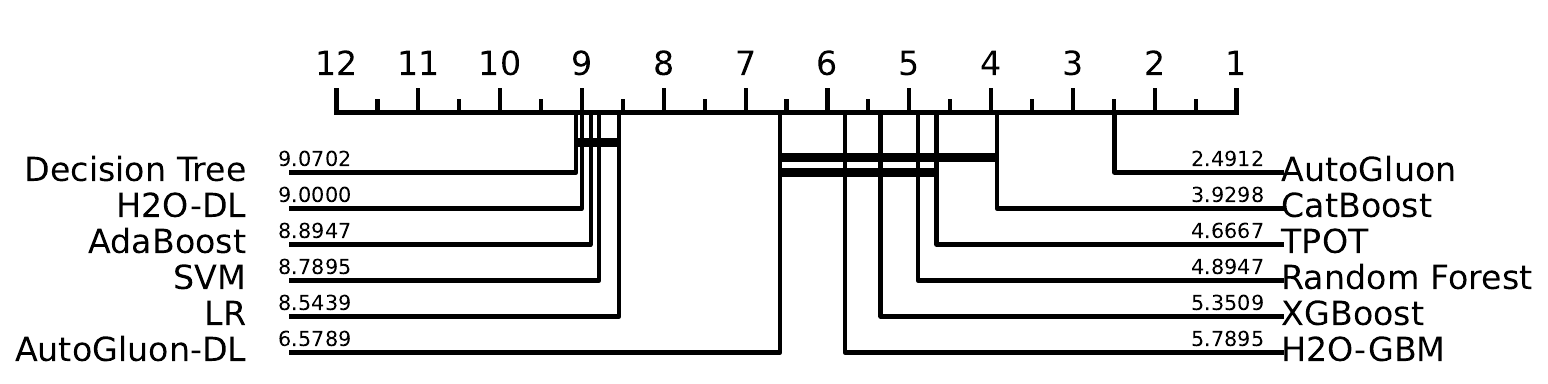}
    \caption{Critical difference diagram for regression tasks based on $R^2$. The best performing model is AutoGluon as higher $R^2$ scores indicate better performance.}
    \label{fig:cd_r2}
\end{figure}

\begin{figure}[!ht]
    \centering
\includegraphics[width=0.92\textwidth]{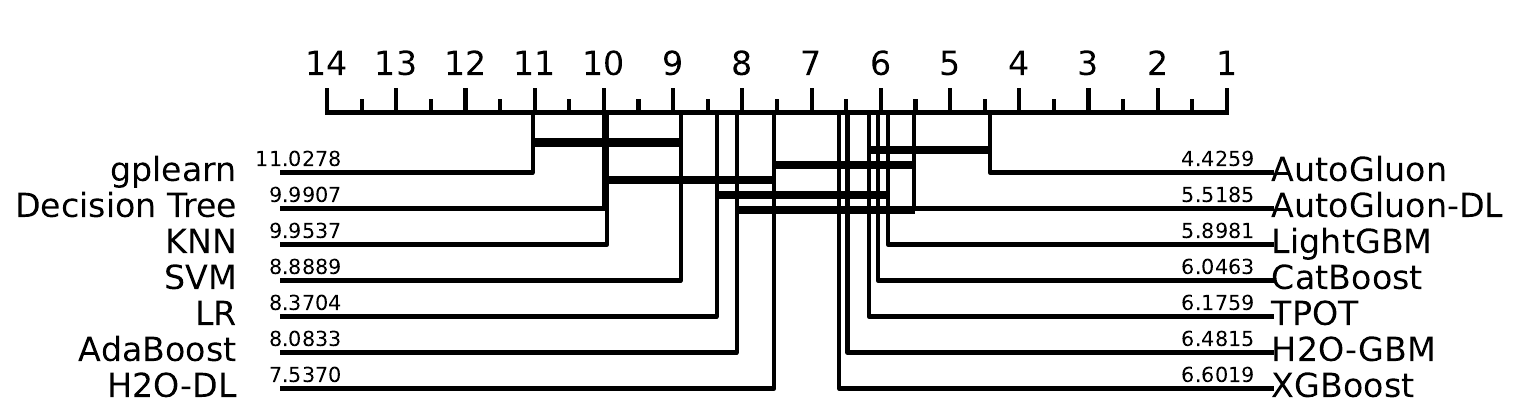}
    \caption{Critical difference diagram for regression tasks based on AUC scores. The best performing model is AutoGluon as higher AUC scores indicate better performance.}
    \label{fig:cd_auc}
\end{figure}

\begin{figure}[!ht]
    \centering
\includegraphics[width=0.92\textwidth]{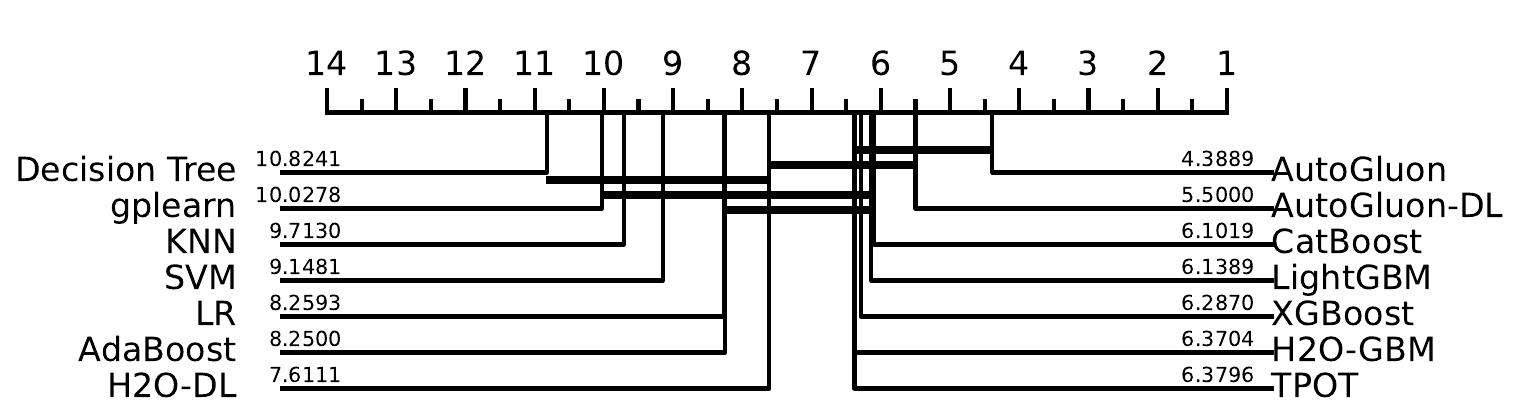}
    \caption{Critical difference diagram for regression tasks based on F1 scores. The best performing model is AutoGluon as higher F1 scores indicate better performance.}
    \label{fig:cd_f1}
\end{figure}

Table \ref{tab:small_all} is equivalent to Table \ref{tab:small_tedl} but includes all 20 models (including those which are not TE or DL based). Tables \ref{tab:lowdim_tedl} and \ref{tab:lowdim_all} present the results for datasets with low dimensions (smaller than 10); Tables \ref{tab:big_tedl} and \ref{tab:big_all} present the results of medium-large datasets (over 10,000 samples); Tables \ref{tab:rmse_tedl} and \ref{tab:rmse_all} present the rankings for the regression datasets based on RMSE; Tables \ref{tab:mae_tedl} and \ref{tab:mae_all} and Tables \ref{tab:r2_tedl} and \ref{tab:r2_all} present the rankings based on MAE and $R^2$, respectively. Finally, the results for the classification datasets based on AUC, accuracy, and F1 scores are summarized in Tables \ref{tab:auc_tedl}, \ref{tab:auc_all}, \ref{tab:acc_tedl}, \ref{tab:acc_all}, \ref{tab:f1_tedl}, and \ref{tab:f1_all}, respectively. Figs. \ref{fig:cd_mae}, \ref{fig:cd_r2}, \ref{fig:cd_auc}, and \ref{fig:cd_f1} present the critical difference diagrams based on additional metrics.

\begin{table}[h!]
\centering
\caption{Performance ranking of all models for small datasets ($<1000$).}
\begin{tabular}{lccccc}
 \hline \hline
Model & Group & \# Best & Average Rank & Median Rank & \# in Top 3 Models \\
\hline \hline
H2O-GBM & TE & 6 & 5.6 & 5 & 12 \\
ResNet & DL & 5 & 6.7 & 7 & 8 \\
CatBoost & TE & 4 & 5.1 & 5 & 14 \\
AutoGluon-DL & DL & 4 & 6.8 & 7.5 & 10 \\
MLP & DL & 4 & 6.3 & 6 & 9 \\
TPOT & TE & 3 & 5.1 & 4 & 15 \\
Random Forest & TE & 3 & 6.1 & 5.5 & 7 \\
LightGBM & TE & 3 & 4.9 & 4 & 13 \\
XGBoost & TE & 2 & 6.6 & 7.5 & 13 \\
H2O-DL & DL & 1 & 9.6 & 11 & 2 \\
DCNV2 & DL & 1 & 9.1 & 10 & 2 \\
AdaBoost & TE & 0 & 8.9 & 8.5 & 2 \\
FT-Transformer & DL & 0 & 10.1 & 11 & 1 \\
TabNet & DL & 0 & 13.9 & 14 & 0 \\
\hline\hline
\end{tabular}
\label{tab:small_all}
\end{table}

\begin{table}[h!]
\centering
\caption{Performance ranking of TE and DL models for datasets with few dimensions ($<10$).}
\begin{tabular}{lccccc}
\hline\hline
Model & Group & \# Best & Average Rank & Median Rank & \# in Top 3 Models \\
\hline\hline
Random Forest & TE & 5 & 6.7 & 6 & 7 \\
H2O-GBM & TE & 4 & 7.1 & 6 & 8 \\
MLP & DL & 4 & 6.3 & 6 & 9 \\
TPOT & TE & 3 & 5.2 & 5 & 13 \\
LightGBM & TE & 3 & 4.9 & 4 & 12 \\
CatBoost & TE & 3 & 5 & 5 & 11 \\
ResNet & DL & 3 & 7.3 & 8 & 7 \\
AutoGluon & DL & 2 & 6.8 & 7 & 9 \\
XGBoost & TE & 1 & 6.6 & 7 & 10 \\
AdaBoost & TE & 1 & 10.9 & 11 & 2 \\
H2O-DL & DL & 1 & 10.6 & 11 & 2 \\
DCNV2 & DL & 1 & 8.2 & 8 & 2 \\
FT-Transformer & DL & 0 & 10.7 & 11 & 1 \\
TabNet & DL & 0 & 12.4 & 13 & 0 \\
\hline\hline
\end{tabular}
\label{tab:lowdim_tedl}
\end{table}

\begin{table}[h!]
\centering
\caption{Performance ranking of all models for datasets with few dimensions ($<10$).}
\begin{tabular}{lccccc}
\hline\hline
Model & Group & \# Best & Average Rank & Median Rank & \# in Top 3 Models \\
\hline\hline
AutoGluon & Other & 11 & 5.3 & 5 & 13 \\
ResNet & DL & 3 & 9 & 10 & 4 \\
H2O-GBM & TE & 3 & 9.3 & 8 & 5 \\
MLP & DL & 2 & 7.7 & 7.5 & 8 \\
LightGBM & TE & 2 & 6.4 & 5 & 9 \\
TPOT & TE & 1 & 6.9 & 7 & 10 \\
XGBoost & TE & 1 & 8.1 & 8 & 6 \\
AutoGluon-DL & DL & 1 & 9.1 & 10 & 5 \\
Decision Tree & Other & 1 & 12.9 & 13.5 & 1 \\
gplearn & Other & 1 & 13.5 & 16 & 2 \\
DCNV2 & DL & 1 & 10.4 & 10 & 2 \\
KNN & Other & 1 & 11.8 & 13 & 5 \\
SVM & Other & 1 & 13.9 & 16 & 1 \\
CatBoost & TE & 1 & 6.4 & 6 & 10 \\
Random Forest & TE & 1 & 8.4 & 7 & 6 \\
H2O-DL & DL & 0 & 13.5 & 15 & 1 \\
AdaBoost & TE & 0 & 13.2 & 15 & 1 \\
FT-Transformer & DL & 0 & 13.9 & 16 & 1 \\
TabNet & DL & 0 & 15.2 & 15 & 0 \\
LR & Other & 0 & 14.1 & 17 & 3 \\
\hline\hline
\end{tabular}
\label{tab:lowdim_all}
\end{table}

\begin{table}[h!]
\centering
\caption{Performance ranking of TE and DL models for medium-large datasets ($>10,000$).}
\begin{tabular}{lccccc}
\hline\hline
Model & Group & \# Best & Average Rank & Median Rank & \# in Top 3 Models \\
\hline\hline
LightGBM & TE & 8 & 4.3 & 3.5 & 19 \\
CatBoost & TE & 8 & 4.3 & 3 & 20 \\
AutoGluon-DL & DL & 5 & 6 & 6 & 10 \\
XGBoost & TE & 3 & 5.5 & 5 & 13 \\
Random Forest & TE & 3 & 6.5 & 7 & 13 \\
H2O-GBM & TE & 3 & 7.4 & 7 & 8 \\
H2O-DL & DL & 2 & 8.1 & 8 & 3 \\
ResNet & DL & 2 & 7.5 & 8 & 6 \\
TPOT & TE & 1 & 6.4 & 6 & 11 \\
MLP & DL & 1 & 8.9 & 9 & 3 \\
AdaBoost & TE & 0 & 11.6 & 12 & 1 \\
FT-Transformer & DL & 0 & 11.2 & 11 & 0 \\
TabNet & DL & 0 & 13.3 & 14 & 0 \\
DCNV2 & DL & 0 & 10.5 & 12 & 1 \\
\hline\hline
\end{tabular}
\label{tab:big_tedl}
\end{table}

\begin{table}[h!]
\centering
\caption{Performance ranking of all models for medium-large datasets ($>10,000$).}
\begin{tabular}{lccccc}
\hline\hline
Model & Group & \# Best & Average Rank & Median Rank & \# in Top 3 Models \\
\hline\hline
AutoGluon & Other & 21 & 3.1 & 1 & 26 \\
AutoGluon-DL & DL & 3 & 7.6 & 7 & 10 \\
ResNet & DL & 2 & 9.6 & 10 & 3 \\
LightGBM & TE & 2 & 6.2 & 4 & 15 \\
SVM & Other & 2 & 14.5 & 16 & 2 \\
TPOT & TE & 1 & 8 & 7 & 5 \\
XGBoost & TE & 1 & 6.9 & 6 & 7 \\
H2O-DL & DL & 1 & 10.7 & 10 & 3 \\
gplearn & Other & 1 & 16.3 & 18 & 2 \\
LR & Other & 1 & 13.2 & 14.5 & 3 \\
H2O & TE & 1 & 8.5 & 8 & 3 \\
KNN & Other & 0 & 11.7 & 12 & 3 \\
Decision Tree & Other & 0 & 12.9 & 14 & 0 \\
CatBoost & TE & 0 & 5.5 & 5 & 12 \\
AdaBoost & TE & 0 & 14.5 & 16 & 0 \\
Random Forest & TE & 0 & 7.5 & 7 & 10 \\
MLP & DL & 0 & 10 & 10.5 & 3 \\
FT-Transformer & DL & 0 & 13.5 & 13 & 0 \\
TabNet & DL & 0 & 17.4 & 17.5 & 0 \\
DCNV2 & DL & 0 & 12.8 & 14 & 1 \\
\hline\hline
\end{tabular}
\label{tab:big_all}
\end{table}

\begin{table}[h!]
\centering
\caption{Performance ranking of TE and DL models for regression datasets, ranked by RMSE score.}
\begin{tabular}{lccccc}
\hline\hline
Model & Group & \# Best & Average Rank & Median Rank & \# in Top 3 Models \\
\hline\hline
CatBoost & TE & 15 & 3.7 & 3 & 31 \\
Random Forest & TE & 9 & 5.5 & 5 & 19 \\
LightGBM & TE & 7 & 4.1 & 3 & 29 \\
AutoGluon-DL & DL & 6 & 7.3 & 7 & 12 \\
TPOT & TE & 4 & 5.3 & 4 & 23 \\
XGBoost & TE & 4 & 5.6 & 5 & 22 \\
H2O-DL & TE & 4 & 6.4 & 6 & 10 \\
ResNet & DL & 4 & 6.9 & 8 & 11 \\
MLP & DL & 3 & 7 & 7 & 10 \\
H2O-DL & DL & 1 & 9.8 & 10.5 & 2 \\
TPOT\_AdaBoost & TE & 0 & 11 & 12 & 2 \\
FT-Transformer & DL & 0 & 10.5 & 11 & 0 \\
TabNet & DL & 0 & 13.8 & 14 & 0 \\
DCNV2 & DL & 0 & 10.8 & 11.5 & 0 \\
\hline\hline
\end{tabular}
\label{tab:rmse_tedl}
\end{table}

\begin{table}[h!]
\centering
\caption{Performance ranking of all models for regression datasets, ranked by RMSE score.}
\begin{tabular}{lccccc}
\hline\hline
Model & Group & \# Best & Average Rank & Median Rank & \# in Top 3 Models \\
\hline\hline
AutoGluon & Other & 30 & 3.5 & 1 & 38 \\
SVM & Other & 4 & 12.9 & 14 & 6 \\
TPOT & TE & 3 & 7 & 5.5 & 11 \\
CatBoost & TE & 3 & 5.2 & 5 & 26 \\
H2O-GBM & TE & 3 & 8.3 & 8 & 7 \\
XGBoost & TE & 3 & 7.5 & 7 & 13 \\
gplearn & Other & 2 & 15.5 & 18 & 3 \\
Random Forest & TE & 2 & 7.2 & 7 & 14 \\
ResNet & DL & 2 & 8.9 & 10 & 3 \\
AutoGluon-DL & DL & 1 & 9.7 & 9 & 8 \\
H2O-DL & DL & 1 & 13 & 13.5 & 2 \\
LR & Other & 1 & 12.9 & 15 & 7 \\
KNN & Other & 1 & 11.1 & 12 & 8 \\
LightGBM & TE & 1 & 5.8 & 4 & 18 \\
Decision Tree & Other & 0 & 13.4 & 14 & 0 \\
AdaBoost & TE & 0 & 13.8 & 14.5 & 0 \\
MLP & DL & 0 & 8.8 & 9 & 7 \\
FT-Transformer & DL & 0 & 13.7 & 14 & 0 \\
TabNet & DL & 0 & 18.2 & 19 & 0 \\
DCNV2 & DL & 0 & 13.7 & 14 & 0 \\
\hline\hline
\end{tabular}
\label{tab:rmse_all}
\end{table}

\begin{table}[h!]
\centering
\caption{Performance ranking of TE and DL models for regression datasets, ranked by MAE score.}
\begin{tabular}{lccccc}
\hline\hline
Model & Group & \# Best & Average Rank & Median Rank & \# in Top 3 Models \\
\hline\hline
CatBoost & TE & 15 & 3.8 & 3 & 31 \\
AutoGluon-DL & DL & 15 & 6.4 & 7 & 18 \\
LightGBM & TE & 6 & 4.2 & 4 & 28 \\
H2O-GBM & TE & 6 & 6.4 & 6 & 11 \\
Random Forest & TE & 5 & 5.2 & 5 & 20 \\
TPOT & TE & 4 & 5.4 & 5 & 22 \\
XGBoost & TE & 3 & 5.8 & 6 & 16 \\
H2O-DL & DL & 1 & 10.3 & 11 & 2 \\
ResNet & DL & 1 & 7.2 & 8 & 12 \\
MLP & DL & 1 & 7.1 & 7 & 9 \\
AdaBoost & TE & 0 & 11.2 & 12 & 1 \\
FT-Transformer & DL & 0 & 10.4 & 11 & 1 \\
TabNet & DL & 0 & 13.8 & 14 & 0 \\
DCNV2 & DL & 0 & 10.6 & 11 & 0 \\
\hline\hline
\end{tabular}
\label{tab:mae_tedl}
\end{table}

\begin{table}[h!]
\centering
\caption{Performance ranking of all models for regression datasets, ranked by MAE score.}
\begin{tabular}{lccccc}
\hline\hline
Model & Group & \# Best & Average Rank & Median Rank & \# in Top 3 Models \\
\hline\hline
AutoGluon & Other & 25 & 3.6 & 2 & 39 \\
CatBoost & TE & 7 & 5.3 & 4.5 & 25 \\
AutoGluon-DL & DL & 6 & 8.1 & 8.5 & 16 \\
TPOT & TE & 3 & 7.2 & 6 & 13 \\
H2O-GBM & TE & 3 & 8.5 & 8 & 6 \\
SVM & Other & 3 & 12.6 & 15 & 5 \\
KNN & Other & 3 & 10.4 & 11 & 8 \\
LightGBM & TE & 2 & 6 & 5 & 17 \\
gplearn & Other & 2 & 15 & 17.5 & 4 \\
ResNet & DL & 1 & 9.1 & 10 & 4 \\
XGBoost & TE & 1 & 7.9 & 8 & 8 \\
H2O-DL & DL & 1 & 13.5 & 14 & 1 \\
TabNet & DL & 0 & 17.9 & 19 & 0 \\
FT-Transformer & DL & 0 & 13.7 & 14 & 0 \\
MLP & DL & 0 & 9.4 & 9 & 5 \\
LR & Other & 0 & 13 & 15 & 6 \\
Decision Tree & Other & 0 & 13.2 & 14 & 1 \\
AdaBoost & TE & 0 & 14.6 & 15 & 0 \\
Random Forest & TE & 0 & 7 & 6 & 13 \\
DCNV2 & DL & 0 & 14 & 14 & 0 \\
\hline\hline
\end{tabular}
\label{tab:mae_all}
\end{table}

\begin{table}[h!]
\centering
\caption{Performance ranking of TE and DL models for regression datasets, ranked by $R^2$ score.}
\begin{tabular}{lccccc}
\hline\hline
Model & Group & \# Best & Average Rank & Median Rank & \# in Top 3 Models \\
\hline\hline
CatBoost & TE & 15 & 3.8 & 3 & 32 \\
Random Forest & TE & 8 & 5.3 & 5 & 19 \\
LightGBM & TE & 8 & 4 & 3 & 30 \\
TPOT & TE & 6 & 5.4 & 4.5 & 19 \\
H2O-GBM & TE & 6 & 6.2 & 6 & 11 \\
AutoGluon-DL & DL & 5 & 7.1 & 7 & 12 \\
MLP & DL & 4 & 6.8 & 7 & 14 \\
XGBoost & TE & 3 & 5.7 & 5.5 & 22 \\
H2O-DL & DL & 1 & 10.2 & 11 & 2 \\
ResNet & DL & 1 & 7 & 8 & 9 \\
AdaBoost & TE & 0 & 10.9 & 11 & 1 \\
FT-Transformer & DL & 0 & 10.6 & 11 & 0 \\
TabNet & DL & 0 & 13.9 & 14 & 0 \\
DCNV2 & DL & 0 & 10.9 & 12 & 0 \\
\hline\hline
\end{tabular}
\label{tab:r2_tedl}
\end{table}

\begin{table}[h!]
\centering
\caption{Performance ranking of all models for regression datasets, ranked by $R^2$ score.}
\begin{tabular}{lccccc}
\hline\hline
Model & Group & \# Best & Average Rank & Median Rank & \# in Top 3 Models \\
\hline\hline
AutoGluon & Other & 31 & 3.3 & 1 & 40 \\
TPOT & TE & 5 & 7.1 & 6 & 10 \\
LightGBM & TE & 3 & 5.5 & 4 & 23 \\
CatBoost & TE & 3 & 5.4 & 4 & 26 \\
H2O-GBM & TE & 3 & 8.2 & 8 & 6 \\
SVM & Other & 3 & 12.6 & 14 & 6 \\
Random Forest & TE & 2 & 7.1 & 6 & 13 \\
AutoGluon-DL & DL & 2 & 9.2 & 9 & 8 \\
XGBoost & TE & 2 & 7.6 & 7 & 12 \\
H2O-DL & DL & 1 & 13.3 & 14 & 2 \\
MLP & DL & 1 & 8.6 & 8 & 8 \\
LR & Other & 1 & 13 & 15 & 6 \\
Decision Tree & Other & 0 & 13.1 & 14 & 0 \\
gplearn & Other & 0 & 16.2 & 18 & 2 \\
KNN & Other & 0 & 11.1 & 12 & 7 \\
ResNet & DL & 0 & 9.2 & 10 & 2 \\
AdaBoost & TE & 0 & 13.7 & 15 & 0 \\
FT-Transformer & DL & 0 & 13.9 & 14 & 0 \\
TabNet & DL & 0 & 18.4 & 19 & 0 \\
DCNV2 & DL & 0 & 13.8 & 14 & 0 \\
\hline\hline
\end{tabular}
\label{tab:r2_all}
\end{table}

\begin{table}[h!]
\centering
\caption{Performance ranking of TE and DL models for classification datasets, ranked by AUC score.}
\begin{tabular}{lccccc}
\hline\hline
Model & Group & \# Best & Average Rank & Median Rank & \# in Top 3 Models \\
\hline\hline
H2O-GBM & TE & 9 & 7.6 & 7 & 18 \\
LightGBM & TE & 8 & 5.8 & 5.5 & 18 \\
ResNet & DL & 6 & 8.1 & 8 & 12 \\
H2O-DL & DL & 5 & 7.8 & 7 & 11 \\
AutoGluon-DL & DL & 5 & 5.9 & 5 & 20 \\
AdaBoost & TE & 4 & 9.2 & 10 & 7 \\
CatBoost & TE & 4 & 6.1 & 5 & 19 \\
DCNV2 & DL & 4 & 7.2 & 7 & 12 \\
TPOT & TE & 3 & 6.6 & 6 & 16 \\
MLP & DL & 3 & 9 & 9 & 7 \\
Random Forest & TE & 2 & 8 & 8 & 9 \\
XGBoost & TE & 1 & 7.3 & 7 & 12 \\
FT-Transformer & DL & 0 & 11.1 & 12 & 1 \\
TabNet & DL & 0 & 12.1 & 13 & 0 \\
\hline\hline
\end{tabular}
\label{tab:auc_tedl}
\end{table}

\begin{table}[h!]
\centering
\caption{Performance ranking of all models for classification datasets, ranked by AUC score.}
\begin{tabular}{lccccc}
\hline\hline
Model & Group & \# Best & Average Rank & Median Rank & \# in Top 3 Models \\
\hline\hline
AutoGluon & Other & 9 & 6.1 & 5 & 20 \\
SVM & Other & 6 & 12 & 14 & 9 \\
ResNet & DL & 5 & 10.5 & 10.5 & 10 \\
LightGBM & TE & 5 & 8 & 7.5 & 15 \\
CatBoost & TE & 4 & 7.9 & 6.5 & 9 \\
AutoGluon-DL & DL & 4 & 7.8 & 7 & 13 \\
DCNV2 & DL & 3 & 9.8 & 10 & 10 \\
H2O-DL & DL & 3 & 9.9 & 10 & 9 \\
MLP & DL & 3 & 10.7 & 11 & 6 \\
H2O-GBM & TE & 3 & 8.9 & 8.5 & 11 \\
TPOT & TE & 2 & 8.4 & 8 & 12 \\
LR & Other & 2 & 10.7 & 10 & 9 \\
gplearn & Other & 1 & 14.4 & 16 & 4 \\
Decision Tree & Other & 1 & 13.3 & 15 & 3 \\
KNN & Other & 1 & 13.2 & 14 & 4 \\
AdaBoost & TE & 1 & 10.7 & 12 & 5 \\
Random Forest & TE & 1 & 9.9 & 10 & 6 \\
XGBoost & TE & 0 & 9.3 & 9 & 6 \\
FT-Transformer & DL & 0 & 14.2 & 15 & 1 \\
TabNet & DL & 0 & 15.9 & 18 & 0 \\
\hline\hline
\end{tabular}
\label{tab:auc_all}
\end{table}

\begin{table}[h!]
\centering
\caption{Performance ranking of TE and DL models for classification datasets, ranked by accuracy score.}
\begin{tabular}{lccccc}
\hline\hline
Model & Group & \# Best & Average Rank & Median Rank & \# in Top 3 Models \\
\hline\hline
LightGBM & TE & 11 & 5.9 & 5.5 & 19 \\
H2O-GBM & TE & 10 & 7.6 & 7 & 21 \\
AutoGluon-DL & DL & 6 & 6.4 & 6 & 15 \\
DCNV2 & DL & 5 & 7.4 & 8 & 9 \\
Random Forest & TE & 4 & 7.4 & 6 & 15 \\
TPOT & TE & 3 & 6.4 & 6 & 15 \\
CatBoost & TE & 3 & 6 & 5 & 22 \\
H2O-DL & DL & 3 & 8.6 & 8 & 7 \\
ResNet & DL & 3 & 7.6 & 8 & 12 \\
XGBoost & TE & 2 & 7.4 & 7 & 9 \\
AdaBoost & TE & 2 & 8.6 & 10 & 9 \\
MLP & DL & 2 & 8.9 & 10 & 8 \\
FT-Transformer & DL & 0 & 11 & 11 & 1 \\
TabNet & DL & 0 & 12.5 & 13 & 0 \\
\hline\hline
\end{tabular}
\label{tab:acc_tedl}
\end{table}

\begin{table}[h!]
\centering
\caption{Performance ranking of all models for classification datasets, ranked by accuracy score}
\begin{tabular}{lccccc}
\hline\hline
Model & Group & \# Best & Average Rank & Median Rank & \# in Top 3 Models \\
\hline\hline
AutoGluon & Other & 13 & 5.7 & 5 & 27 \\
LightGBM & TE & 6 & 7.9 & 7 & 15 \\
SVM & Other & 5 & 11.6 & 12 & 6 \\
H2O-GBM & TE & 5 & 8.2 & 8 & 17 \\
DCNV2 & DL & 4 & 10.1 & 10 & 8 \\
AutoGluon-DL & DL & 4 & 8.7 & 7 & 11 \\
ResNet & DL & 2 & 9.3 & 9.5 & 7 \\
H2O-DL & DL & 2 & 10.7 & 11 & 4 \\
TPOT & TE & 2 & 8.4 & 7 & 11 \\
CatBoost & TE & 2 & 7.7 & 6 & 9 \\
Random Forest & TE & 2 & 9 & 7 & 9 \\
LR & Other & 2 & 11.2 & 11.5 & 10 \\
KNN & Other & 1 & 12.5 & 13 & 3 \\
XGBoost & TE & 1 & 9.7 & 9 & 6 \\
Decision Tree & Other & 1 & 13.6 & 15 & 3 \\
AdaBoost & TE & 1 & 10.9 & 13 & 7 \\
MLP & DL & 1 & 11.3 & 11.5 & 7 \\
gplearn & Other & 0 & 15 & 16 & 1 \\
FT-Transformer & DL & 0 & 13.9 & 15 & 1 \\
TabNet & DL & 0 & 16.3 & 18 & 0 \\
\hline\hline
\end{tabular}
\label{tab:acc_all}
\end{table}

\begin{table}[h!]
\centering
\caption{Performance ranking of TE and DL models for classification datasets, ranked by F1 score.}
\begin{tabular}{lccccc}
\hline\hline
Model & Group & \# Best & Average Rank & Median Rank & \# in Top 3 Models \\
\hline\hline
H2O-GBM & TE & 8 & 7.6 & 7 & 17 \\
DCNV2 & DL & 8 & 6.9 & 7 & 13 \\
LightGBM & TE & 7 & 5.7 & 6 & 17 \\
AutoGluon-DL & DL & 7 & 6 & 4 & 21 \\
H2O-DL & DL & 6 & 8 & 7 & 10 \\
AdaBoost & TE & 4 & 9.1 & 10 & 10 \\
CatBoost & TE & 4 & 6.2 & 5 & 22 \\
TPOT & TE & 3 & 6.5 & 6 & 13 \\
XGBoost & TE & 2 & 6.9 & 7 & 11 \\
Random Forest & TE & 2 & 7.1 & 7 & 13 \\
ResNet & DL & 2 & 9.3 & 9 & 7 \\
MLP & DL & 1 & 9.5 & 10 & 5 \\
FT-Transformer & DL & 0 & 11.1 & 12 & 1 \\
TabNet & DL & 0 & 11.7 & 13 & 2 \\
\hline\hline
\end{tabular}
\label{tab:f1_tedl}
\end{table}

\begin{table}[h!]
\centering
\caption{Performance ranking of all models for classification datasets, ranked by F1 score.}
\begin{tabular}{lccccc}
\toprule
Model & Group & \# Best & Average Rank & Median Rank & \# in Top 3 Models \\
\midrule
AutoGluon & Other & 11 & 5.6 & 5 & 23 \\
DCNV2 & DL & 7 & 9.4 & 9 & 12 \\
SVM & Other & 5 & 12 & 13.5 & 9 \\
AutoGluon-DL & DL & 4 & 7.8 & 6 & 16 \\
AdaBoost & TE & 4 & 11.3 & 13 & 7 \\
H2O-DL & DL & 4 & 9.9 & 10 & 9 \\
LightGBM & TE & 3 & 8.1 & 7 & 12 \\
H2O-GBM & TE & 3 & 8.8 & 9 & 11 \\
gplearn & Other & 2 & 12.9 & 14 & 8 \\
TPOT & TE & 2 & 8.4 & 8 & 10 \\
CatBoost & TE & 2 & 7.6 & 6 & 10 \\
LR & Other & 2 & 10.7 & 10 & 9 \\
XGBoost & TE & 1 & 9.2 & 9 & 5 \\
KNN & Other & 1 & 12.8 & 13 & 4 \\
Random Forest & TE & 1 & 8.9 & 9 & 7 \\
ResNet & DL & 1 & 11.7 & 12 & 5 \\
MLP & DL & 1 & 12.3 & 13 & 4 \\
Decision Tree & Other & 0 & 14.3 & 15 & 0 \\
FT-Transformer & DL & 0 & 14.3 & 15 & 1 \\
TabNet & DL & 0 & 15.3 & 16.5 & 0 \\
\bottomrule
\end{tabular}
\label{tab:f1_all}
\end{table}
\clearpage

\subsection{Computer resources}
We ran the experiments with 15 Google Colab sessions, using the high-RAM (51 GB) configuration with CPU. The total computing time including initial attempts and robustness tests  is estimated at 4,000 hours.

\end{document}